\documentclass{bmvc2k}

\usepackage{multirow}
\usepackage{amssymb}
\usepackage[misc,geometry]{ifsym}

\title{PaRK-Detect: Towards Efficient Multi-Task \\ Satellite Imagery Road Extraction via \\ Patch-Wise Keypoints Detection}

\addauthor{Shenwei Xie}{xieshenwei@bupt.edu.cn}{1}
\addauthor{Wanfeng Zheng}{zhengwanfeng@bupt.edu.cn}{1}
\addauthor{Zhenglin Xian}{2019213264@bupt.edu.cn}{1}
\addauthor{Junli Yang}{yangjunli@bupt.edu.cn}{1}
\addauthor{Chuang Zhang}{zhangchuang@bupt.edu.cn}{1}
\addauthor{Ming Wu \Letter}{wuming@bupt.edu.cn}{1}

\addinstitution{
 Beijing University of Posts and\\
 Telecommunications, China
}

\runninghead{Xie, Zheng, Xian, Yang, Zhang, Wu}{PaRK-Detect}

\def\etal{\emph{et al}\bmvaOneDot}

\begin{document}

\maketitle

\begin{abstract}
Automatically extracting roads from satellite imagery is a fundamental yet challenging computer vision task in the field of remote sensing. Pixel-wise semantic segmentation-based approaches and graph-based approaches are two prevailing schemes. However, prior works show the imperfections that semantic segmentation-based approaches yield road graphs with low connectivity, while graph-based methods with iterative exploring paradigms and smaller receptive fields focus more on local information and are also time-consuming. In this paper, we propose a new scheme for multi-task satellite imagery road extraction, Patch-wise Road Keypoints Detection (PaRK-Detect). Building on top of D-LinkNet architecture and adopting the structure of keypoint detection, our framework predicts the position of patch-wise road keypoints and the adjacent relationships between them to construct road graphs in a single pass. Meanwhile, the multi-task framework also performs pixel-wise semantic segmentation and generates road segmentation masks. We evaluate our approach against the existing state-of-the-art methods on DeepGlobe, Massachusetts Roads, and RoadTracer datasets and achieve competitive or better results. We also demonstrate a considerable outperformance in terms of inference speed.

\end{abstract}

\section{Introduction}
\label{sec:intro}
\begin{figure}
\includegraphics[width=13cm]{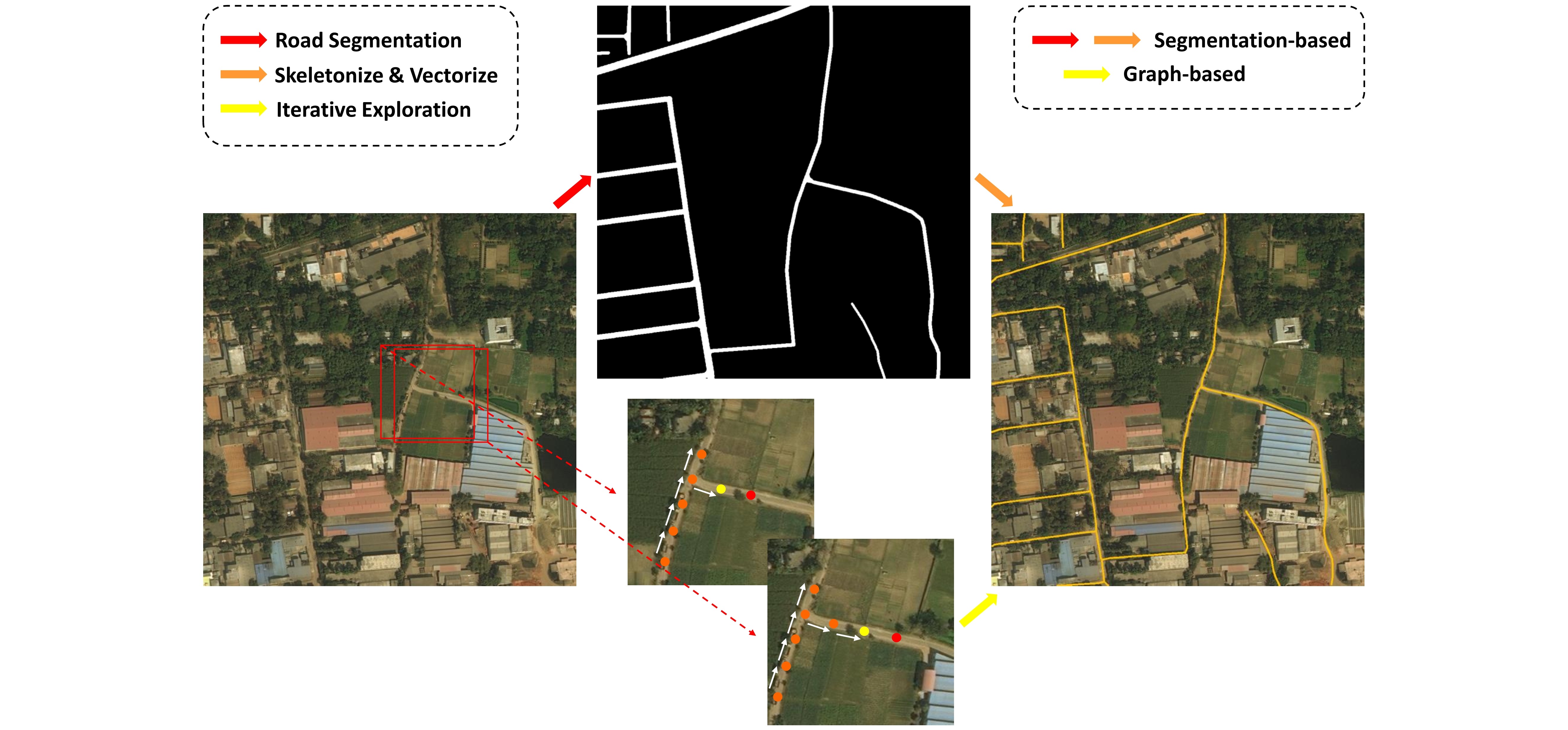}
\caption{Illustration of segmentation-based approaches and graph-based approaches for road extraction from satellite imagery. Red and orange arrows present segmentation-based approaches that perform road segmentation with skeletonization and vectorization. The yellow arrow stands for the iterative exploration utilized in graph-based approaches.}
\label{fig1}
\end{figure}

Road extraction from satellite imagery is a long-standing computer vision 
task and a hot research topic in the field of remote sensing. Extracting 
roads is challenging due to the complexity of road networks along with the 
diversity of geographical environments. With the rapid development of image 
interpretation and the cutting-edge architecture of Convolutional Neural 
Networks (CNNs) designed for computer vision tasks, automatically 
extracting wide-coverage road graphs from high-resolution aerial images 
becomes more accurate and efficient. As shown in Fig. \ref{fig1}, prior prevailing 
works fall into two main categories: pixel-wise semantic segmentation-based 
approaches and graph-based approaches. Semantic segmentation-based approaches 
perform binary classification to distinguish road pixels from background pixels 
and form segmentation maps as an intermediate representation of road 
graphs. Road networks are constructed by road centerlines and edges through 
skeletonizing segmentation masks and edge detection. In contrast, 
graph-based approaches extract road graphs directly by yielding vertices 
and edges of road networks. The iterative exploration paradigm has been 
utilized to search for the next move in correspondence to the roads 
in satellite imagery.

However, extensive experiments reveal that these two approaches both 
have some drawbacks when it comes to certain evaluation perspectives. 
Under the circumstances of severe occlusion and geographical environments 
with high complexity, segmentation-based approaches usually yield road 
graphs with low connectivity. On the other hand, graph-based approaches 
benefit from iterative exploring paradigms that overcome connectivity 
issues, nevertheless, leading to smaller receptive fields and focusing 
more on local features without a global perspective. Moreover, the 
iterative exploring paradigm only searches one step at a time which is 
time-consuming for road graph inference.

According to the above discussions, we propose a new scheme for satellite imagery road extraction, patch-wise road keypoints detection (termed as PaRK-Detect), which is inspired by keypoint detection and combines the inherent superiority in the global receptive field of semantic 
segmentation-based approaches while ravels out its drawbacks that the absence of connectivity supervision affects road network topology. More specifically, by dividing high-resolution satellite imagery into small patches, the road graphs are partitioned section by section. The framework built on top of D-LinkNet architecture determines whether each patch contains a road section and if so, then detects the road keypoint in that patch. Establishing the adjacent relationships which represent the connection status between the road patches is also conducted by the framework. In this way, our proposed approaches enable road graph construction in a single pass rather than iterative exploration, in other words, a much faster inference speed becomes reality. Meanwhile, our proposed multi-task network also performs pixel-wise road semantic segmentation to enrich contextual information and enhance information capture capabilities.

Through evaluating our approach against the existing state-of-the-art methods 
on the publicly available DeepGlobe dataset, Massachusetts Roads dataset, and 
RoadTracer dataset, we show that our new scheme achieves competitive or better results 
based on several metrics for road extraction and yields road networks with high 
accuracy and topological integrity.

Our main contributions can be concluded as follows:
\begin{itemize}
\item We propose a patch-wise road keypoints detection scheme that combines 
road extraction with keypoint detection to construct road graphs in a single 
pass.
\item We propose an efficient multi-task framework that performs road 
segmentation and road graphs extraction simultaneously.
\item We conduct extensive experiments over widely-used datasets and achieve 
competitive or better results using several road extraction metrics.
\end{itemize}

\section{Related Work}
{\bf Pixel-wise Semantic Segmentation-based Approaches.}  With the increasing 
popularity and dominance of convolutional neural network (CNN) architecture, 
utilizing the symmetrical~\cite{ronneberger2015u,chaurasia2017linknet} 
or heterogeneous~\cite{long2015fully,zhao2017pyramid,chen2017rethinking} 
encoder-decoder framework becomes the favorite paradigm for semantic segmentation 
tasks. Road segmentation is a binary classification task that splits pixels of 
remote sensing images into road and non-road. Many refined and improved 
segmentation networks~\cite{buslaev2018fully,zhang2018road,zhou2018d,mattyus2017deeproadmapper} 
have been proposed for this scene. 
For example, Buslaev \etal~\cite{buslaev2018fully} proposed a fully convolutional 
network (FCN) that consists of pretrained ResNet~\cite{he2016deep} and a decoder 
adapted from vanilla U-Net~\cite{ronneberger2015u}. 
Zhang \etal~\cite{zhang2018road} proposed ResUnet, an elegant architecture with 
better performance that combines the strengths of residual learning and U-Net. 
Zhou \etal~\cite{zhou2018d} proposed D-LinkNet that built with LinkNet~\cite{chaurasia2017linknet} architecture and adopts dilated convolution layers in its central part to enlarge 
the receptive field and ensemble multi-scale features. 
DeepRoadMapper~\cite{mattyus2017deeproadmapper} improves the loss function and 
the post-processing strategy that reasons about missing connections in the 
extracted road topology as the shortest-path problem. 
Although these powerful networks provide a global perspective of road 
networks and detailed road features, segmentation-based approaches suffer 
from the absence of connectivity supervision and yield segmentation masks 
with redundant road information which is meaningless for road graphs construction.
Our proposed patch-wise road keypoints detection scheme fully considers the 
connectivity status and focuses on the road graph while retaining a global perspective.
\\ \hspace*{\fill} \\
\noindent {\bf Graph-based Approaches.} Unlike semantic segmentation approaches, 
graph-based approaches directly extract vertices and edges to construct road graphs. 
Through exploiting the iterative exploring paradigm, the next moves 
searched along the road are successively added to the existed road graph. 
For instance, RoadTracer~\cite{bastani2018roadtracer} utilizes the iterative searching 
guided by a CNN-based decision function that decides the stop probability 
and angle vector of the next move, while VecRoad~\cite{tan2020vecroad} proposes a 
point-based iterative exploration scheme with segmentation-cues guidance and 
flexible steps. 
Apart from these iterative methods, He \etal~\cite{he2020sat2graph} proposed 
Sat2Graph with a graph-tensor encoding (GTE) scheme which encodes the road graph 
into a tensor representation. Bahl \etal~\cite{bahl2021road} combined an FCN in 
charge of locating road intersections, dead ends, and turns, and a Graph Neural 
Network (GNN) that predicts links between these points. Our scheme constructs 
road graphs in a single pass through detecting patch-wise road keypoints without 
the iterative exploring paradigm.
\\ \hspace*{\fill} \\
\noindent {\bf Multi-Task Road Extraction Approaches.} Road extraction can have 
different purposes and tasks in different scenarios. City planning and road network 
updating focus on topology information in road graphs, while emergency response 
may focus on road damage status that can be evaluated through the segmentation mask. 
Cheng \etal~\cite{cheng2017automatic} proposed CasNet, a cascaded end-to-end CNN, 
to simultaneously cope with road detection and centerline extraction. 
Yang \etal~\cite{yang2019road} proposed a recurrent convolution neural network 
U-Net (RCNN-UNet) with the multitask learning scheme. 
Wei \etal~\cite{wei2020simultaneous} extracted road surface and centerline concurrently 
and proposed a multistage framework that consists of boosting segmentation, multiple 
starting points tracing, and fusion. 
Our proposed framework conducts road segmentation and road graph construction 
simultaneously to improve accuracy, robustness, and efficiency.
\\ \hspace*{\fill} \\
\noindent {\bf Keypoint Detection.}  In recent years, keypoint detection has enjoyed 
substantial attention and has a promising application prospect in automatic driving 
and human tracking. Human pose estimation, facial landmark detection, and detection 
of specific types of object parts are collectively known as keypoint detection. 
Since the pioneering work DeepPose~\cite{toshev2014deeppose}, many DNN-based methods 
for keypoint detection have been proposed. 
Basically, constructing ground truth falls into three categories: directly using 
coordinates of keypoints~\cite{toshev2014deeppose}, constructing  heatmap~\cite{newell2016stacked,cai2020learning}, and jointly utilizing heatmap and  offsets~\cite{shi2019improved}. 
The stacked hourglass network (SHN)~\cite{newell2016stacked} has been widely used as its 
repeated bottom-up, top-down processing with successive pooling and upsampling ensures 
different features are extracted from feature maps of different scales to achieve the 
best performance. 
Cai \etal~\cite{cai2020learning} proposed Residual Steps Network (RSN) that aggregates 
intra-level features to retain rich low-level spatial information. Additionally, 
Pose Refine Machine (PRM), an efficient attention mechanism, has been proposed 
to make a trade-off between the local and global representation for further refinement. 
Shi \etal~\cite{shi2019improved} proposed a novel improved SHN with the multi-scale 
feature aggregation module designed for accurate and robust facial landmark detection. 
Then, an offset learning sub-network is adopted to refine the inferred keypoints. 
Inspired by these works, our proposed approach performs patch-wise offset learning for 
road keypoints.

\section{Methods}
This section presents our proposed patch-wise road keypoints detection scheme. Section \ref{scheme} defines fundamental elements of road status used in patch-wise road representation and illustrates the patch-wise road keypoints detection scheme in detail. Once the definitions of road representation and the procedures of the scheme are learned, the architecture of the proposed multi-task framework is explained in section \ref{framework}. We also describe novel learning objectives designed for our scheme in section \ref{loss function}. Finally, an effective graph optimization strategy is briefly illustrated in section \ref{post process}.

\begin{figure}
\centering
\includegraphics[width=13cm]{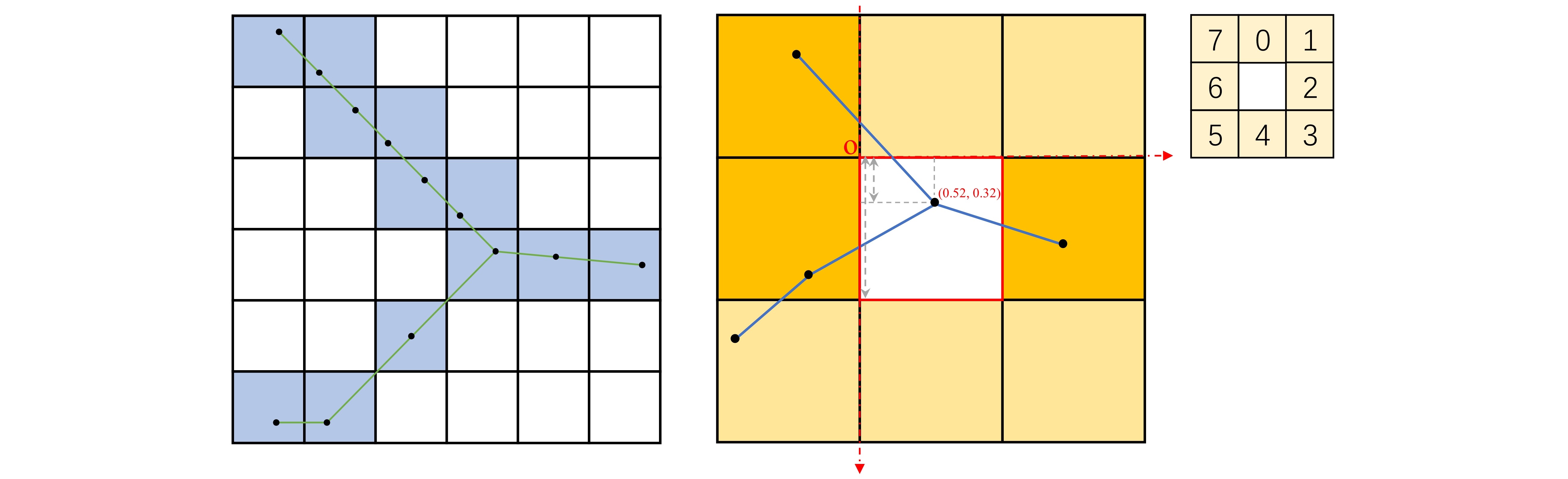}
\caption{{\bf Illustration of PaRK-Detect scheme. Left:} blue patches contain road while white patches are non-road, black dots are road keypoints, and green lines represent links. {\bf Right:} The reference point of relative offset is the upper left corner of a patch. Dark yellow patches are linked with the center patch while light yellow ones are not. We order the eight adjacent patches into numbers 0-7. Here the linked patches are 2, 6, and 7.}
\label{fig2}
\end{figure}

\subsection{PaRK-Detect Scheme}
\label{scheme}

As introduced above, our PaRK-Detect scheme detects patch-wise road keypoints and constructs road graphs in a single pass. A high-resolution satellite image (e.g. of size 1024$\times$1024 pixels) can be divided into non-overlapping patches (e.g. $64^{2}$ patches of size 16$\times$16 pixels) as the minimum unit for road representation. Fig. \ref{fig2} shows the illustration of the PaRK-Detect scheme. Three fundamental elements are defined for road representation in each patch: $P$, $S$, and $L$.

After dividing entire large satellite imagery into $N^{2}$ non-overlapping small patches, the road networks are also partitioned section by section by the grids. $P$ stands for the probability of whether a patch contains a road section. If a patch does not cover road, then we should not consider it any further.

For patches that cover road, $S$ stands for the position of patch-wise road keypoint, which is also defined as the most crucial point in a patch. The priority of road intersection is higher than the road endpoint. If there is no intersection or endpoint in a patch that contains a road section, then the keypoint is defined at the geometric center of the road segment in that patch, which is also called the midpoint. We use relative offset to represent the position of the patch-wise road keypoint. 

Furthermore, $L$ stands for the link status or connectivity status between these patches. Each patch can be linked to eight patches around itself. If two adjacent patches are linked, then it means there exists a road segment between the road keypoints of them. We use eight probability values to represent link status. A road graph consists of vertexes and edges which are formed by these keypoints and links.

\subsection{Framework Architecture}
\label{framework}

\begin{figure}
\includegraphics[width=13cm]{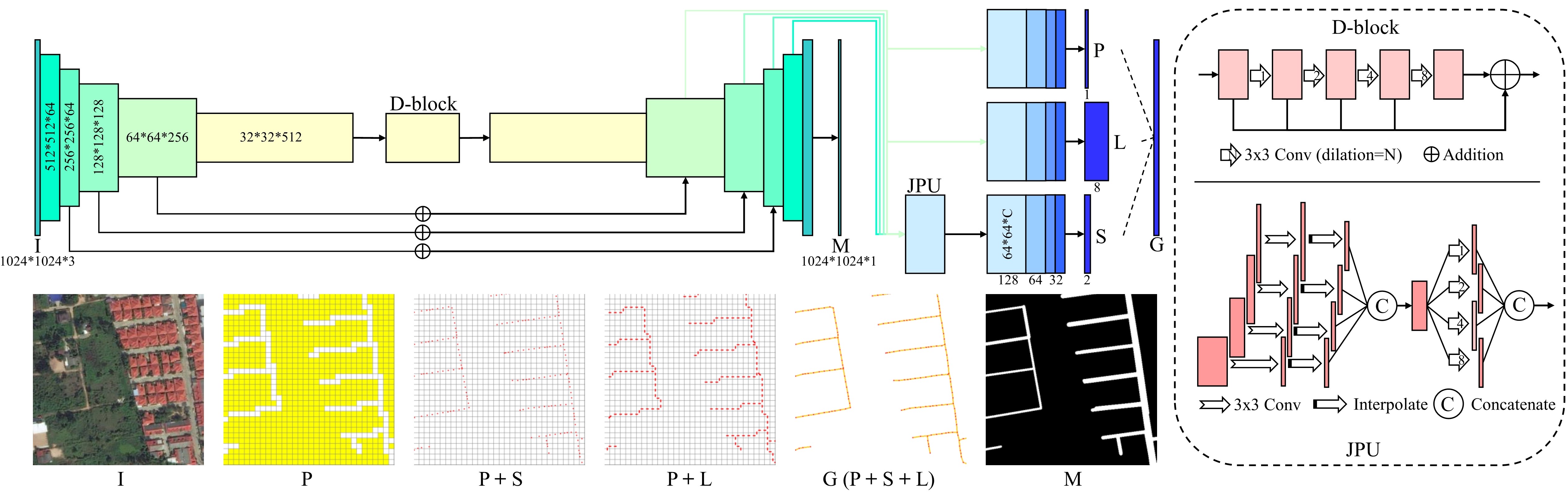}
\caption{{\bf Overview of our proposed multi-task framework architecture.} The rectangles are feature maps of different scales. {\bf I:} input satellite image, {\bf P:} patch-wise road probability, yellow patches represent non-road while white patches represent road, {\bf S:} patch-wise road keypoint position, {\bf L:} patch-wise link status, {\bf G:} road graph, {\bf M:} road segmentation mask. Here we just show $32^{2}$ patches out of $64^{2}$ for better presentation.}
\label{fig3}
\end{figure}

As illustrated by Fig. \ref{fig3}, we designed a multi-task framework for our PaRK-Detect scheme that performs road segmentation and road graph construction simultaneously. Following the encoder-decoder frameworks paradigm, we use ResNet34~\cite{he2016deep} pretrained on ImageNet~\cite{deng2009imagenet} dataset as encoders to extract high-level semantic information such as road features (e.g. the encoder part downsamples satellite images of size 1024$\times$1024 and outputs feature maps of size 32$\times$32). We also adopt dilated convolution with skip connections same as the bottleneck of D-LinkNet~\cite{zhou2018d} after encoders to increase the receptive field of feature points while keeping detailed information.

For road segmentation, the framework retains the decoder that utilizes the transposed convolution and the skip connection same as LinkNet~\cite{chaurasia2017linknet}, restoring the resolution of feature map (e.g. from 32$\times$32 to 1024$\times$1024). 

For road graph construction, three networks are designed for predicting $P$, $L$, and $S$ respectively. These three fundamental elements of patch-wise road representation are indispensable yet sufficient to construct road graphs. Patch-wise probability predicting network performs classification of whether a patch covers road. Patch-wise link predicting network determines the connection status between a patch and its eight adjacent patches. Patch-wise position predicting network locates the patch-wise keypoint in each patch. Both probability and link predicting networks only consist of convolutional layers and batch normalization with feature maps of the same size (e.g. 64$\times$64) and decreasing channels. In order to keep low-level detailed information, position predicting network combines joint pyramid upsampling (JPU)~\cite{wu2019fastfcn} additionally for multi-scale feature fusion. The outputs of these three networks are feature maps (e.g. of size 64$\times$64) with 1, 8, and 2 channels respectively.

\subsection{Learning Objectives}
\label{loss function}

To train our multi-task framework, we define a joint learning objective for road segmentation and road graph construction. The joint learning objective can be formulated as follow: 

$$\mathcal{L} = \mathcal{L}_{seg} + \mathcal{L}_{graph}$$

Road segmentation loss remains the same as segmentation-based methods~\cite{zhou2018d} that evaluate the difference between labels and predicted masks with binary cross entropy (BCE) and dice coefficient loss.
For road graph construction, we define the training loss functions $\mathcal{L}_{P}$, $\mathcal{L}_{L}$, and $\mathcal{L}_{S}$ for $P$, $L$, and $S$ respectively as follows: 
$$\mathcal{L}_{graph} = \alpha\mathcal{L}_{P} + \beta\mathcal{L}_{S} + \gamma\mathcal{L}_{L}$$
$$\mathcal{L}_{P}= -\sum_{i=1}^{N^{2}}\left[P_{gt}^{i}\log{P_{pre}^{i}}+\left(1-P_{gt}^{i}\right)\log{\left(1-P_{pre}^{i}\right)}\right]$$
$$\mathcal{L}_{L}= -\frac{1}{\Omega_{P}}\sum_{\Omega_{P}}\sum_{j=1}^{8}\left[L_{gt}^{ij}\log{L_{pre}^{ij}}+\left(1-L_{gt}^{ij}\right)\log{\left(1-L_{pre}^{ij}\right)}\right]$$
$$\mathcal{L}_{S}= -\frac{1}{\Omega_{P}}\sum_{\Omega_{P}}\sum_{j=1}^{2}\left|S_{gt}^{ij}-S_{pre}^{ij}\right|$$
where $\Omega_{P}=\left\{i|P_{gt}^{i}=1,i\in\left[1,N^{2}\right]\right\}$, which is a set of patches that contain road, and j stands for the dimension of each patch-wise element. Patch-wise road probability loss and patch-wise link status loss use binary cross entropy (BCE) as the loss function while patch-wise road keypoint position loss uses mean absolute error (MAE). On account that we only consider patches that cover road and ignore other patches when it comes to locating road keypoints and evaluating connection status, we define partial loss functions for L and S. $\alpha$, $\beta$, and $\gamma$ are the weights of the loss functions for three patch-wise elements.

\begin{figure}
\includegraphics[width=6.5cm]{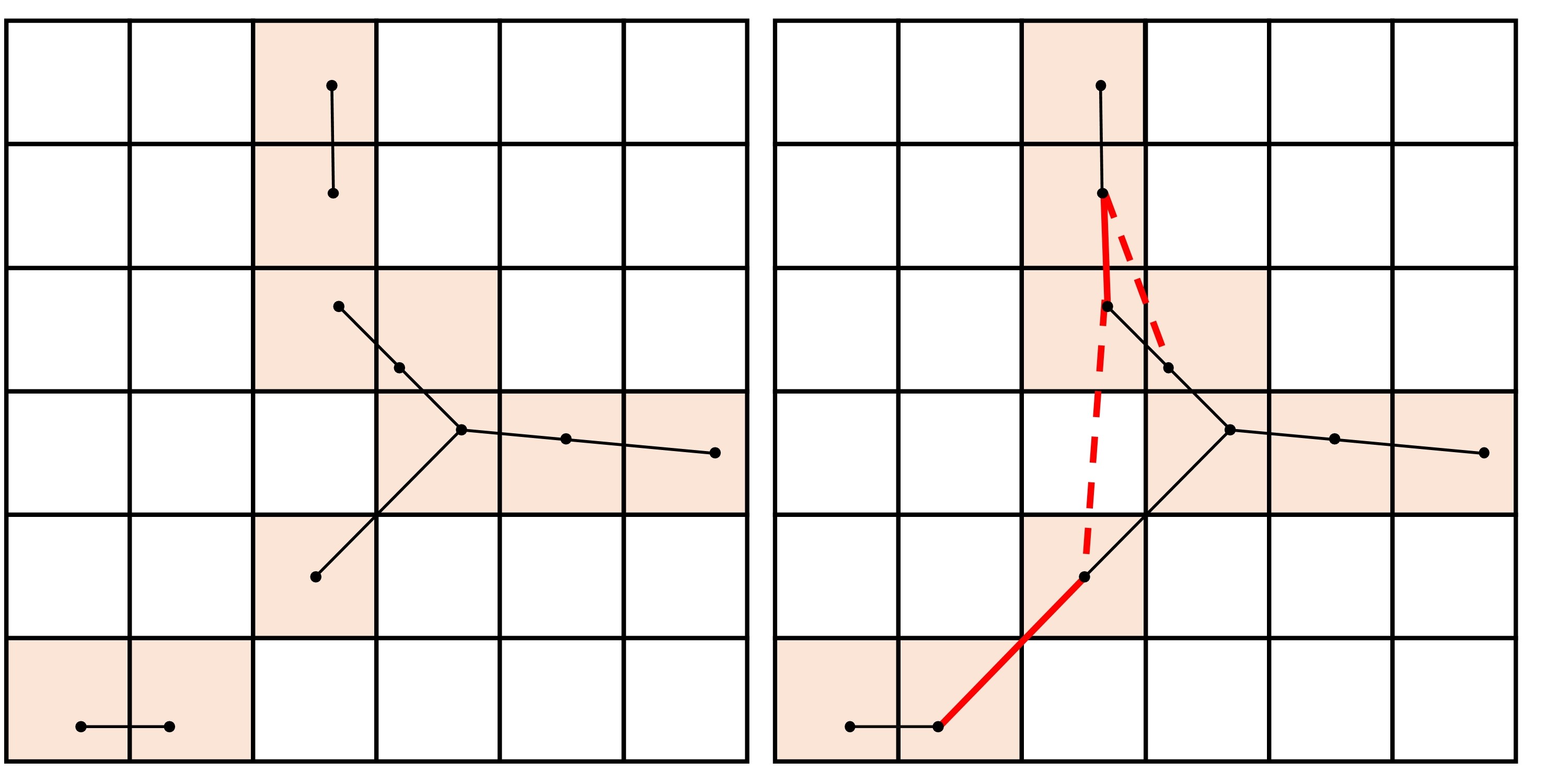}
\includegraphics[width=6.5cm]{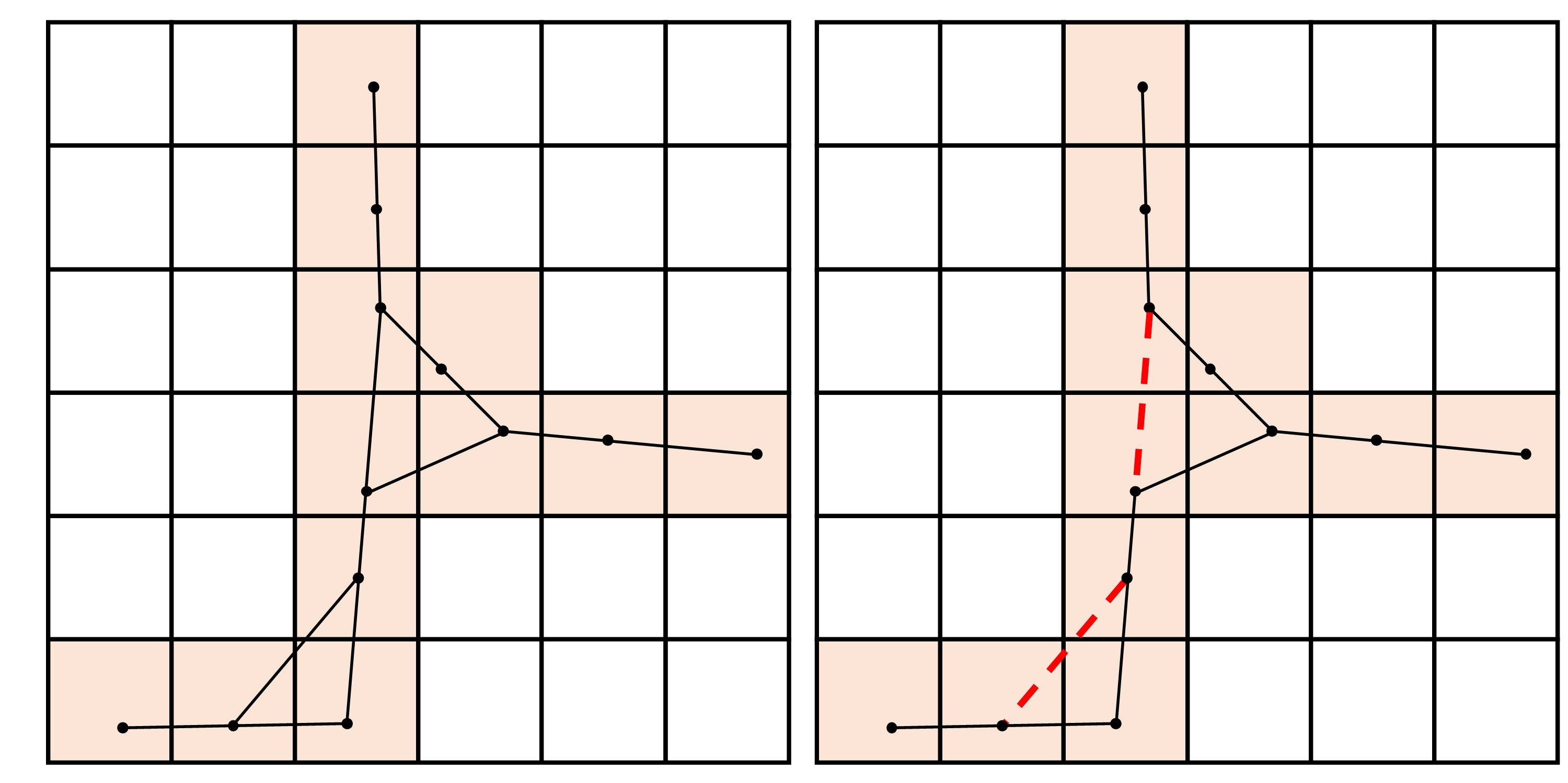}
\caption{{\bf Illustration of graph optimization strategies. Left:} connecting adjacent but unconnected endpoints. Red solid lines are links added while red dotted lines are links that should not be added. {\bf Right:} removing triangle and quadrilateral. Red dotted lines are links removed.}
\label{fig4}
\end{figure}

\subsection{Graph Optimization}
\label{post process}
As shown in Fig. \ref{fig4}, we develop several graph optimization strategies to refine the patch-wise link status to improve the performance of road network construction. 

{\bf Connecting adjacent but unconnected endpoints.} Due to the occlusion of buildings and trees, the connectivity status of adjacent patches not always appears realistic. Through extensive observation, we find out that adjacent but unconnected endpoints predicted by the framework are mostly connected actually.

{\bf Removing triangle and quadrilateral.} Several intersections are so large that cover several patches, this will leads to unnecessary links between these patches and inaccurate road graphs. We remove the diagonal link if a triangular connection between adjacent patches appears. In addition, we remove the longest link if a quadrilateral connection between adjacent patches appears.

\section{Experiments}

\subsection{Experimental Datasets}
{\bf DeepGlobe Road Extraction Dataset}~\cite{demir2018deepglobe} is collected from the DigitalGlobe platform and other datasets, and consists of high-resolution satellite images of size 1024$\times$1024 pixels. It covers images captured over Thailand, Indonesia, and India, and the ground resolution is 0.5m/pixel. We utilize the original training set which consists of 6226 images and randomly split it into 3984, 997, and 1245 images for training, validation, and testing respectively.

{\bf Massachusetts Roads Dataset}~\cite{mnih2013machine} is collected from publicly available imagery and metadata and consists of satellite images of size 1500$\times$1500 pixels. It covers images captured over a wide variety of urban, suburban, and rural regions of the state of Massachusetts, and the ground resolution is 1m/pixel. We crop the 1171 original images into 3168 images of size 1024$\times$1024 pixels and randomly split them into 2027, 507, and 634 images.

{\bf RoadTracer Dataset}~\cite{bastani2018roadtracer} is collected from Google and consists of high-resolution satellite images of size 4096$\times$4096 pixels. It covers the urban core of forty cities across six countries, and the ground resolution is 0.6m/pixel. We crop and resize the training set which consists of 300 images into 1200 images of size 1024$\times$1024 pixels and randomly split it into 960 and 240 images for training and validation. 15 8192$\times$8192 images are used for testing. 

The split of the datasets aims for approximate distribution of 64$\%$, 16$\%$, and 20$\%$.

\subsection{Implementation Details}
All the experiments are conducted with the PyTorch framework on NVIDIA 3090Ti GPUs. 
We use DeepGlobe Road Extraction Dataset~\cite{demir2018deepglobe}, Massachusetts Roads Dataset~\cite{mnih2013machine}, and RoadTracer Dataset~\cite{bastani2018roadtracer} for experiments. The input satellite images are with a dimension of 1024$\times$1024 and the patch size is set to 16$\times$16. During training, the batch size is fixed as 4 and the total epoch is set to 300. We use Adam optimizer with the learning rate originally set to 2e-4 and reduced by 5 times once the loss keeps relatively higher for several epochs. The weights of loss functions $\alpha$, $\beta$, and $\gamma$ are set to 0.5, 1, and 1 respectively. To avoid overfitting, we utilize a variety of data augmentation ways, including horizontal flip, vertical flip, random rotation, and color jittering. We use the pixel-based F1 score~\cite{van2018spacenet} and APLS~\cite{van2018spacenet} metric to evaluate road graphs while using the IoU score to evaluate segmentation performance.

\subsection{Comparison with Other Methods}
For segmentation-based approaches, we compare the PaRK-Detect scheme with D-LinkNet~\cite{zhou2018d} on DeepGlobe and Massachusetts Roads Datasets. The comparison results are shown in Tab.\;\ref{Table1}. Note that road graphs can be constructed through skeletonizing segmentation masks generated by segmentation-based approaches for evaluating APLS metric. Only focusing on pixel-wise semantic features, D-LinkNet acquires more road morphological information as well as contour features, nevertheless, this is redundant for constructing road graphs. On the contrary, our approach provides connectivity supervision while retaining the global receptive field and achieves outstanding performance with pixel-based F1 score and APLS metric. Besides, as shown in Tab.\;\ref{inferspeed}, our method outperforms D-LinkNet by 23.1$\%$ in terms of inference speed as segmentation-based approaches require skeletonization and vectorization in post-processing procedures.

For graph-based approaches, we compare our scheme with RoadTracer~\cite{bastani2018roadtracer} and VecRoad~\cite{tan2020vecroad} on RoadTracer Dataset. The comparison results are shown in Tab.\;\ref{Table2}. Our approach achieves competitive results compared with RoadTracer and VecRoad while the inference speed of the PaRK-detect scheme far exceeds that of these approaches. 

\begin{table}
	\begin{center}
	\scriptsize
	\begin{tabular}{|c|c|c|c|c|c|c|}
        \hline 
        \multirow{2}{*}{Method} & \multicolumn{3}{c|}{DeepGlobe} & \multicolumn{3}{c|}{Massachusetts Roads}\\ 
        \cline{2-7} 
        & P-F1 & APLS & IoU & P-F1 & APLS & IoU\\ 
        \hline
        \hline
        D-LinkNet & 74.87 & 62.74 & 65.27 & 71.10 & 59.08 & 56.41\\
        \hline
        \hline
        {\bf Ours} & {\bf 78.04} & {\bf 69.83} & {\bf 65.53} & {\bf 74.80} & {\bf 63.37} & {\bf 56.80}\\
        \hline
	\end{tabular} 
	\end{center}
	\vspace{-4pt}
	\caption{Comparison with segmentation-based approach on DeepGlobe and Massachusetts Roads Dataset.}
	\label{Table1}
\end{table}

\begin{table}[!t]
	\centering
	\begin{minipage}{0.44\textwidth}
		\centering
		\scriptsize
		\begin{tabular}{|c|c|c|}
            \hline 
            Method & P-F1 & APLS\\ 
            \hline
            \hline
            VecRoad & 72.56 & {\bf 64.59}\\
            RoadTracer & 55.81 & 45.09\\
            \hline
            \hline
            {\bf Ours} & {\bf 74.50} & 63.10\\
            \hline
		\end{tabular}
		\vspace{6pt}
		\caption{Comparison with graph-based approaches on RoadTracer Dataset.}
		\label{Table2}
	\end{minipage}\quad
	\begin{minipage}{0.46\textwidth}
		\centering
		\scriptsize
		\begin{tabular}{|c|c|c|}
            \hline
            Method & Type & 8192$\times$8192 \\
            \hline
            \hline 
            VecRoad & Graph & 1327.7 \\
            RoadTracer & Graph & 343.4 \\
            D-LinkNet & Seg & 10.1 \\ 
            \hline
            \hline
            {\bf Ours} & PaRK-Detect & {\bf 7.7} \\
            \hline
        \end{tabular}
		\vspace{6pt}
		\caption{Run-time in seconds of different approaches on one 8192$\times$8192 test image.}
		\label{inferspeed}
	\end{minipage}
\end{table}

\subsection{Ablation Studies}
\begin{figure}
\includegraphics[width=6.5cm]{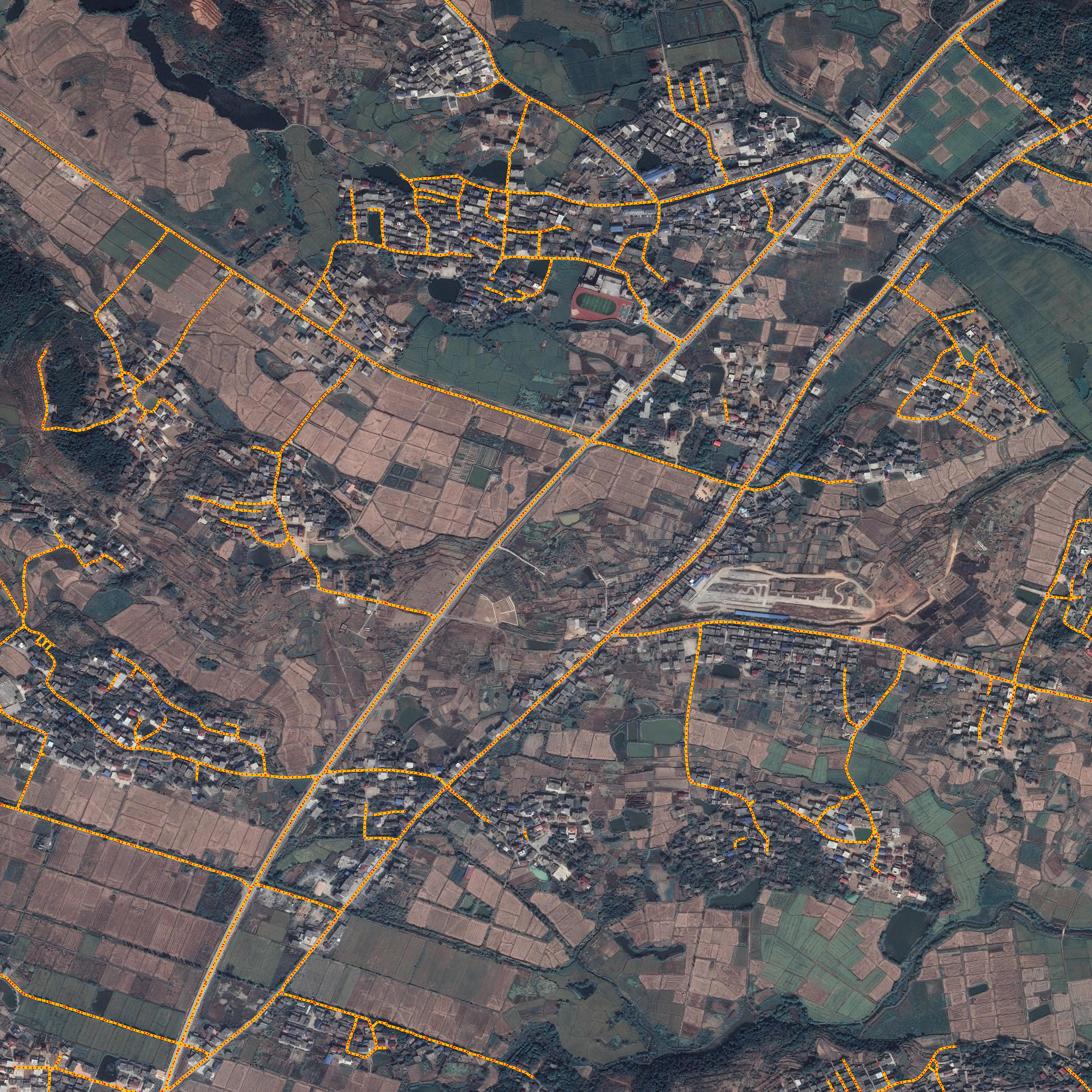}
\includegraphics[width=6.5cm]{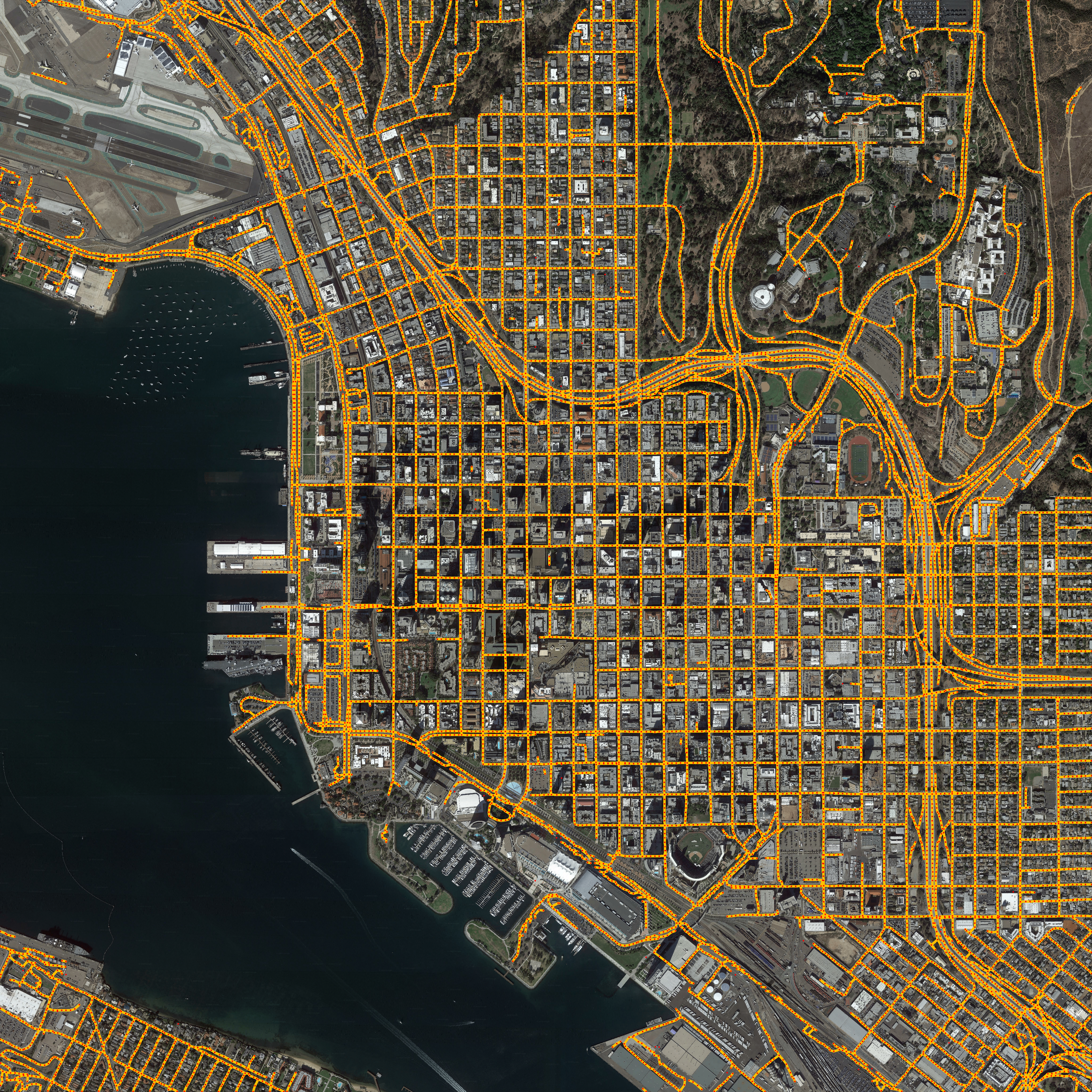}
\caption{Results of constructed road graphs with PaRK-Detect scheme.}
\label{fig5}
\end{figure}

Fig. \ref{fig5} presents the results of constructed road graphs based on our proposed PaRK-Detect scheme. To verify the effectiveness of our approach and the framework architecture, we conduct ablations studies for multi-scale feature fusion and graph optimization strategies. As shown in Tab. \ref{ablation}, introducing JPU to perform multi-scale feature fusion improves the overall performance of our method. Graph optimization strategies improve the pixel-based F1 score and APLS of constructed road graphs by 1.7$\%$ and 9.0$\%$ respectively on the DeepGlobe dataset and by 0.6$\%$ and 10.8$\%$ respectively on the Massachusetts Roads dataset. Performing graph optimization will not influence the IoU metric of road segmentation.

Besides, in order to explore the impact of the multi-task paradigm for road extraction, we remove the decoder part of our framework and compare the road graph construction results with the original framework. As shown in Table 4, simply performing road graph construction without road segmentation diminishes pixel-based F1 score and APLS, which implies road segmentation can be served as additional supervision and provides road morphological information and contour features for achieving better alignment of roads.

Patch size is a crucial hyperparameter in our PaRK-Detect scheme and different patch sizes lead to different framework architectures and discrepant road alignment accuracy. We use 16$\times$16 as the default setting and conduct extensive experiments to explore the impact of patch size. We observe that the appropriate patch size is determined by the width of the road in satellite imagery. If the road width is mostly concentrated between 12-24 pixels (e.g. DeepGlobe and Massachusetts Roads dataset), 16$\times$16 is the suitable patch size. If the road width is mostly 24-48 pixels (e.g. RoadTracer dataset), we should choose 32$\times$32. However, we utilize interpolation to resize the RoadTracer dataset and the patch size remains 16$\times$16. 

\newcommand{\tabincell}[2]{\begin{tabular}{@{}#1@{}}#2\end{tabular}}
\begin{table}[!t]
\scriptsize
\begin{center}
\begin{tabular}{|c c c c|c|c|c|c|c|c|}
    \hline
    \multirow{2}{*}{\tabincell{c}{multi-scale\\feature fusion}} & \multirow{2}{*}{\tabincell{c}{graph\\optimization}} & 
    \multirow{2}{*}{segmentation} & 
    \multirow{2}{*}{\tabincell{c}{patch\\size}} 
    & \multicolumn{3}{c|}{DeepGlobe} & \multicolumn{3}{c|}{Massachusetts Roads} \\
    \cline{5-10}
    & ~ & ~ & ~ & P-F1 & APLS & IoU & P-F1 & APLS & IoU\\
    \hline
    \hline
    ~ & \checkmark & \checkmark & 16$\times$16 & 77.44 & 69.07 & 65.24 & 74.36 & 62.80 & 56.43\\
    \checkmark & ~ & \checkmark & 16$\times$16 & 76.70 & 64.06 & {\bf 65.53} & 74.35 & 57.17 & {\bf 56.80}\\
    \checkmark & \checkmark & ~ & 16$\times$16 & 76.97 & 69.64 &   -   & 72.98 & 58.07 &   -  \\
    \checkmark & \checkmark & \checkmark & 16$\times$16 & {\bf 78.04} & {\bf 69.83} & {\bf 65.53} & {\bf 74.80} & {\bf 63.37} & {\bf 56.80}\\
    \checkmark & \checkmark & \checkmark & 8$\times$8 & 66.48 & 58.12 & 65.49 & 65.52 & 49.68 & 56.72\\
    \checkmark & \checkmark & \checkmark & 32$\times$32 & 76.41 & 67.91 & 65.48 & 73.57 & 60.90 & 56.74\\
    \hline
\end{tabular}
\end{center}
\vspace{6pt}
\caption{Ablation studies of our proposed scheme. No mask will be predicted for evaluating IoU if we remove the road segmentation part of the proposed multi-task framework.}
\label{ablation}
\end{table}

\section{Conclusion}
In this paper, we propose a patch-wise road keypoints detection scheme that combines road extraction with keypoint detection to construct road graphs in a single pass. We also propose an efficient multi-task framework that performs road segmentation and road graph construction simultaneously. We conduct extensive experiments over widely-used datasets and achieve competitive or better road extraction results with superior inference speed against existing works. Through ablation experiments, we verify the effectiveness of our approach and demonstrate the impact of the multi-task paradigm and patch size. We believe our work presents an important step forward and we hope this paper can inspire future works on road extraction from satellite imagery.

\section{Acknowledgement}
This work was supported in part by the National Natural Science Foundation of China under Grant 62076093 and the High-Performance Computing Platform of BUPT.

\bibliography{PaRK-Detect}

\begin{thebibliography}{26}
\providecommand{\natexlab}[1]{#1}
\providecommand{\url}[1]{\texttt{#1}}
\expandafter\ifx\csname urlstyle\endcsname\relax
  \providecommand{\doi}[1]{doi: #1}\else
  \providecommand{\doi}{doi: \begingroup \urlstyle{rm}\Url}\fi

\bibitem[Bahl et~al.(2021)Bahl, Bahri, and Lafarge]{bahl2021road}
Gaetan Bahl, Mehdi Bahri, and Florent Lafarge.
\newblock Road extraction from overhead images with graph neural networks.
\newblock \emph{arXiv preprint arXiv:2112.05215}, 2021.

\bibitem[Bastani et~al.(2018)Bastani, He, Abbar, Alizadeh, Balakrishnan,
  Chawla, Madden, and DeWitt]{bastani2018roadtracer}
Favyen Bastani, Songtao He, Sofiane Abbar, Mohammad Alizadeh, Hari
  Balakrishnan, Sanjay Chawla, Sam Madden, and David DeWitt.
\newblock Roadtracer: Automatic extraction of road networks from aerial images.
\newblock In \emph{Proceedings of the IEEE Conference on Computer Vision and
  Pattern Recognition}, pages 4720--4728, 2018.

\bibitem[Buslaev et~al.(2018)Buslaev, Seferbekov, Iglovikov, and
  Shvets]{buslaev2018fully}
Alexander Buslaev, Selim Seferbekov, Vladimir Iglovikov, and Alexey Shvets.
\newblock Fully convolutional network for automatic road extraction from
  satellite imagery.
\newblock In \emph{Proceedings of the IEEE conference on computer vision and
  pattern recognition workshops}, pages 207--210, 2018.

\bibitem[Cai et~al.(2020)Cai, Wang, Luo, Yin, Du, Wang, Zhang, Zhou, Zhou, and
  Sun]{cai2020learning}
Yuanhao Cai, Zhicheng Wang, Zhengxiong Luo, Binyi Yin, Angang Du, Haoqian Wang,
  Xiangyu Zhang, Xinyu Zhou, Erjin Zhou, and Jian Sun.
\newblock Learning delicate local representations for multi-person pose
  estimation.
\newblock In \emph{European Conference on Computer Vision}, pages 455--472.
  Springer, 2020.

\bibitem[Chaurasia and Culurciello(2017)]{chaurasia2017linknet}
Abhishek Chaurasia and Eugenio Culurciello.
\newblock Linknet: Exploiting encoder representations for efficient semantic
  segmentation.
\newblock In \emph{2017 IEEE Visual Communications and Image Processing
  (VCIP)}, pages 1--4. IEEE, 2017.

\bibitem[Chen et~al.(2017)Chen, Papandreou, Schroff, and
  Adam]{chen2017rethinking}
Liang-Chieh Chen, George Papandreou, Florian Schroff, and Hartwig Adam.
\newblock Rethinking atrous convolution for semantic image segmentation.
\newblock \emph{arXiv preprint arXiv:1706.05587}, 2017.

\bibitem[Cheng et~al.(2017)Cheng, Wang, Xu, Wang, Xiang, and
  Pan]{cheng2017automatic}
Guangliang Cheng, Ying Wang, Shibiao Xu, Hongzhen Wang, Shiming Xiang, and
  Chunhong Pan.
\newblock Automatic road detection and centerline extraction via cascaded
  end-to-end convolutional neural network.
\newblock \emph{IEEE Transactions on Geoscience and Remote Sensing},
  55\penalty0 (6):\penalty0 3322--3337, 2017.

\bibitem[Demir et~al.(2018)Demir, Koperski, Lindenbaum, Pang, Huang, Basu,
  Hughes, Tuia, and Raskar]{demir2018deepglobe}
Ilke Demir, Krzysztof Koperski, David Lindenbaum, Guan Pang, Jing Huang, Saikat
  Basu, Forest Hughes, Devis Tuia, and Ramesh Raskar.
\newblock Deepglobe 2018: A challenge to parse the earth through satellite
  images.
\newblock In \emph{Proceedings of the IEEE Conference on Computer Vision and
  Pattern Recognition Workshops}, pages 172--181, 2018.

\bibitem[Deng et~al.(2009)Deng, Dong, Socher, Li, Li, and
  Fei-Fei]{deng2009imagenet}
Jia Deng, Wei Dong, Richard Socher, Li-Jia Li, Kai Li, and Li~Fei-Fei.
\newblock Imagenet: A large-scale hierarchical image database.
\newblock In \emph{2009 IEEE conference on computer vision and pattern
  recognition}, pages 248--255. Ieee, 2009.

\bibitem[He et~al.(2016)He, Zhang, Ren, and Sun]{he2016deep}
Kaiming He, Xiangyu Zhang, Shaoqing Ren, and Jian Sun.
\newblock Deep residual learning for image recognition.
\newblock In \emph{Proceedings of the IEEE conference on computer vision and
  pattern recognition}, pages 770--778, 2016.

\bibitem[He et~al.(2020)He, Bastani, Jagwani, Alizadeh, Balakrishnan, Chawla,
  Elshrif, Madden, and Sadeghi]{he2020sat2graph}
Songtao He, Favyen Bastani, Satvat Jagwani, Mohammad Alizadeh, Hari
  Balakrishnan, Sanjay Chawla, Mohamed~M Elshrif, Samuel Madden, and
  Mohammad~Amin Sadeghi.
\newblock Sat2graph: Road graph extraction through graph-tensor encoding.
\newblock In \emph{European Conference on Computer Vision}, pages 51--67.
  Springer, 2020.

\bibitem[Long et~al.(2015)Long, Shelhamer, and Darrell]{long2015fully}
Jonathan Long, Evan Shelhamer, and Trevor Darrell.
\newblock Fully convolutional networks for semantic segmentation.
\newblock In \emph{Proceedings of the IEEE conference on computer vision and
  pattern recognition}, pages 3431--3440, 2015.

\bibitem[M{\'a}ttyus et~al.(2017)M{\'a}ttyus, Luo, and
  Urtasun]{mattyus2017deeproadmapper}
Gell{\'e}rt M{\'a}ttyus, Wenjie Luo, and Raquel Urtasun.
\newblock Deeproadmapper: Extracting road topology from aerial images.
\newblock In \emph{Proceedings of the IEEE international conference on computer
  vision}, pages 3438--3446, 2017.

\bibitem[Mnih(2013)]{mnih2013machine}
Volodymyr Mnih.
\newblock \emph{Machine learning for aerial image labeling}.
\newblock University of Toronto (Canada), 2013.

\bibitem[Newell et~al.(2016)Newell, Yang, and Deng]{newell2016stacked}
Alejandro Newell, Kaiyu Yang, and Jia Deng.
\newblock Stacked hourglass networks for human pose estimation.
\newblock In \emph{European conference on computer vision}, pages 483--499.
  Springer, 2016.

\bibitem[Ronneberger et~al.(2015)Ronneberger, Fischer, and
  Brox]{ronneberger2015u}
Olaf Ronneberger, Philipp Fischer, and Thomas Brox.
\newblock U-net: Convolutional networks for biomedical image segmentation.
\newblock In \emph{International Conference on Medical image computing and
  computer-assisted intervention}, pages 234--241. Springer, 2015.

\bibitem[Shi and Wang(2019)]{shi2019improved}
Husen Shi and Zengfu Wang.
\newblock Improved stacked hourglass network with offset learning for robust
  facial landmark detection.
\newblock In \emph{2019 9th International Conference on Information Science and
  Technology (ICIST)}, pages 58--64. IEEE, 2019.

\bibitem[Tan et~al.(2020)Tan, Gao, Li, Cheng, and Ren]{tan2020vecroad}
Yong-Qiang Tan, Shang-Hua Gao, Xuan-Yi Li, Ming-Ming Cheng, and Bo~Ren.
\newblock Vecroad: Point-based iterative graph exploration for road graphs
  extraction.
\newblock In \emph{Proceedings of the IEEE/CVF Conference on Computer Vision
  and Pattern Recognition}, pages 8910--8918, 2020.

\bibitem[Toshev and Szegedy(2014)]{toshev2014deeppose}
Alexander Toshev and Christian Szegedy.
\newblock Deeppose: Human pose estimation via deep neural networks.
\newblock In \emph{Proceedings of the IEEE conference on computer vision and
  pattern recognition}, pages 1653--1660, 2014.

\bibitem[Van~Etten et~al.(2018)Van~Etten, Lindenbaum, and
  Bacastow]{van2018spacenet}
Adam Van~Etten, Dave Lindenbaum, and Todd~M Bacastow.
\newblock Spacenet: A remote sensing dataset and challenge series.
\newblock \emph{arXiv preprint arXiv:1807.01232}, 2018.

\bibitem[Wei et~al.(2020)Wei, Zhang, and Ji]{wei2020simultaneous}
Yao Wei, Kai Zhang, and Shunping Ji.
\newblock Simultaneous road surface and centerline extraction from large-scale
  remote sensing images using cnn-based segmentation and tracing.
\newblock \emph{IEEE Transactions on Geoscience and Remote Sensing},
  58\penalty0 (12):\penalty0 8919--8931, 2020.

\bibitem[Wu et~al.(2019)Wu, Zhang, Huang, Liang, and Yu]{wu2019fastfcn}
Huikai Wu, Junge Zhang, Kaiqi Huang, Kongming Liang, and Yizhou Yu.
\newblock Fastfcn: Rethinking dilated convolution in the backbone for semantic
  segmentation.
\newblock \emph{arXiv preprint arXiv:1903.11816}, 2019.

\bibitem[Yang et~al.(2019)Yang, Li, Ye, Lau, Zhang, and Huang]{yang2019road}
Xiaofei Yang, Xutao Li, Yunming Ye, Raymond~YK Lau, Xiaofeng Zhang, and Xiaohui
  Huang.
\newblock Road detection and centerline extraction via deep recurrent
  convolutional neural network u-net.
\newblock \emph{IEEE Transactions on Geoscience and Remote Sensing},
  57\penalty0 (9):\penalty0 7209--7220, 2019.

\bibitem[Zhang et~al.(2018)Zhang, Liu, and Wang]{zhang2018road}
Zhengxin Zhang, Qingjie Liu, and Yunhong Wang.
\newblock Road extraction by deep residual u-net.
\newblock \emph{IEEE Geoscience and Remote Sensing Letters}, 15\penalty0
  (5):\penalty0 749--753, 2018.

\bibitem[Zhao et~al.(2017)Zhao, Shi, Qi, Wang, and Jia]{zhao2017pyramid}
Hengshuang Zhao, Jianping Shi, Xiaojuan Qi, Xiaogang Wang, and Jiaya Jia.
\newblock Pyramid scene parsing network.
\newblock In \emph{Proceedings of the IEEE conference on computer vision and
  pattern recognition}, pages 2881--2890, 2017.

\bibitem[Zhou et~al.(2018)Zhou, Zhang, and Wu]{zhou2018d}
Lichen Zhou, Chuang Zhang, and Ming Wu.
\newblock D-linknet: Linknet with pretrained encoder and dilated convolution
  for high resolution satellite imagery road extraction.
\newblock In \emph{Proceedings of the IEEE Conference on Computer Vision and
  Pattern Recognition Workshops}, pages 182--186, 2018.

\end{thebibliography}
\end{document}